\documentclass{article}

\usepackage[nonatbib, preprint]{neurips_data_2023}

\usepackage[utf8]{inputenc} 
\usepackage[T1]{fontenc}    
\usepackage{hyperref}       
\usepackage{url}            
\usepackage{booktabs}       
\usepackage{amsfonts}  
\usepackage{amsmath}
\usepackage{nicefrac}       
\usepackage{microtype}      
\usepackage{xcolor}         
\usepackage{biblatex}
\usepackage{pifont}
\usepackage{graphicx}
\usepackage{comment}
\usepackage{xspace}
\usepackage{lipsum}

\usepackage{soul}

\let\svthefootnote\thefootnote
\newcommand\freefootnote[1]{%
  \let\thefootnote\relax%
  \footnotetext{#1}%
  \let\thefootnote\svthefootnote%
}

\bibliography{egbib.bib}
\usepackage{multirow}
\newcommand{\xmark}{\ding{55}}%
\newcommand{\datasetname}{{ReutersViLNews}\xspace}
\title{Multi-modal News Understanding with Professionally Labelled Videos (\datasetname)}

\author{%
  Shih-Han Chou$^{* 1,2}$ \\
  \And 
  Matthew Kowal$^{* 1,4,7}$ \\
  \And
  Yasmin Niknam$^{* 1,3}$ \\
  \AND 
  Diana Moyano$^{1}$ \\
  \And
  Shayaan Mehdi$^{1}$ \\
  \And
  Richard Pito$^{5}$ \\
  \And
  Cheng Zhang$^{5}$ \\
  \And
  Ian Knopke$^{5}$ \\
  \And
  Sedef Akinli Kocak$^{1}$ \\
  \And
  Leonid Sigal$^{1,2,6}$ \\
  \And
  Yalda Mohsenzadeh$^{1,3}$ \\
  \AND
  \normalfont Vector Institute$^{1}$, University of British Columbia$^{2}$, Western University$^{3}$ \\ \vspace{-1cm}
  \AND 
  \normalfont York University$^{4}$, Reuters News Agency$^{5}$, CIFAR$^{6}$, Toyota Research Institute$^{7}$ \\
}

\begin{document}

\maketitle

\vspace{-0.8cm}



\begin{abstract}
While progress has been made in the domain of video-language understanding, current state-of-the-art algorithms are still limited in their ability to understand videos at high levels of abstraction, such as news-oriented videos. On the other hand, humans easily amalgamate information from video and language to infer information beyond what is visually observable in the pixels. An example of this is watching a news story, where the context of the event can play as big of a role in understanding the story as the event itself. Towards a solution for designing this ability in algorithms, here we present a large-scale analysis on an in-house dataset collected by the Reuters News Agency, called Reuters Video-Language News shortened to "\datasetname" dataset which focuses on high-level video-language understanding with an emphasis on long-form news. The \datasetname Dataset consists of long-form news videos collected and labeled by professionals in the news industry over several years and contains prominent news reporting from around the world. Each video involves a single story and contains action shots of the actual event, interviews with people associated with the event, footage from nearby areas, and more. \datasetname dataset contains videos from seven subject categories: disaster, finance, entertainment, health, politics, sports, and miscellaneous with annotations from high-level to low-level, title caption, visual video description, high-level story description, keywords, and location. We first present a detailed analysis of the dataset statistics of \datasetname compared to previous datasets. Then we benchmark state-of-the-art approaches for four different video-language tasks. The results suggest that news-oriented videos are a substantial challenge for current video-language understanding algorithms and we conclude by providing future directions in designing approaches to solve the \datasetname dataset. 
\end{abstract} \vspace{-0.5cm}
\def\thefootnote{*}\footnotetext{Equal Contribution}
\section{Introduction}
\begin{figure}[!t]
    \centering    
    \includegraphics[width=1.0\textwidth]{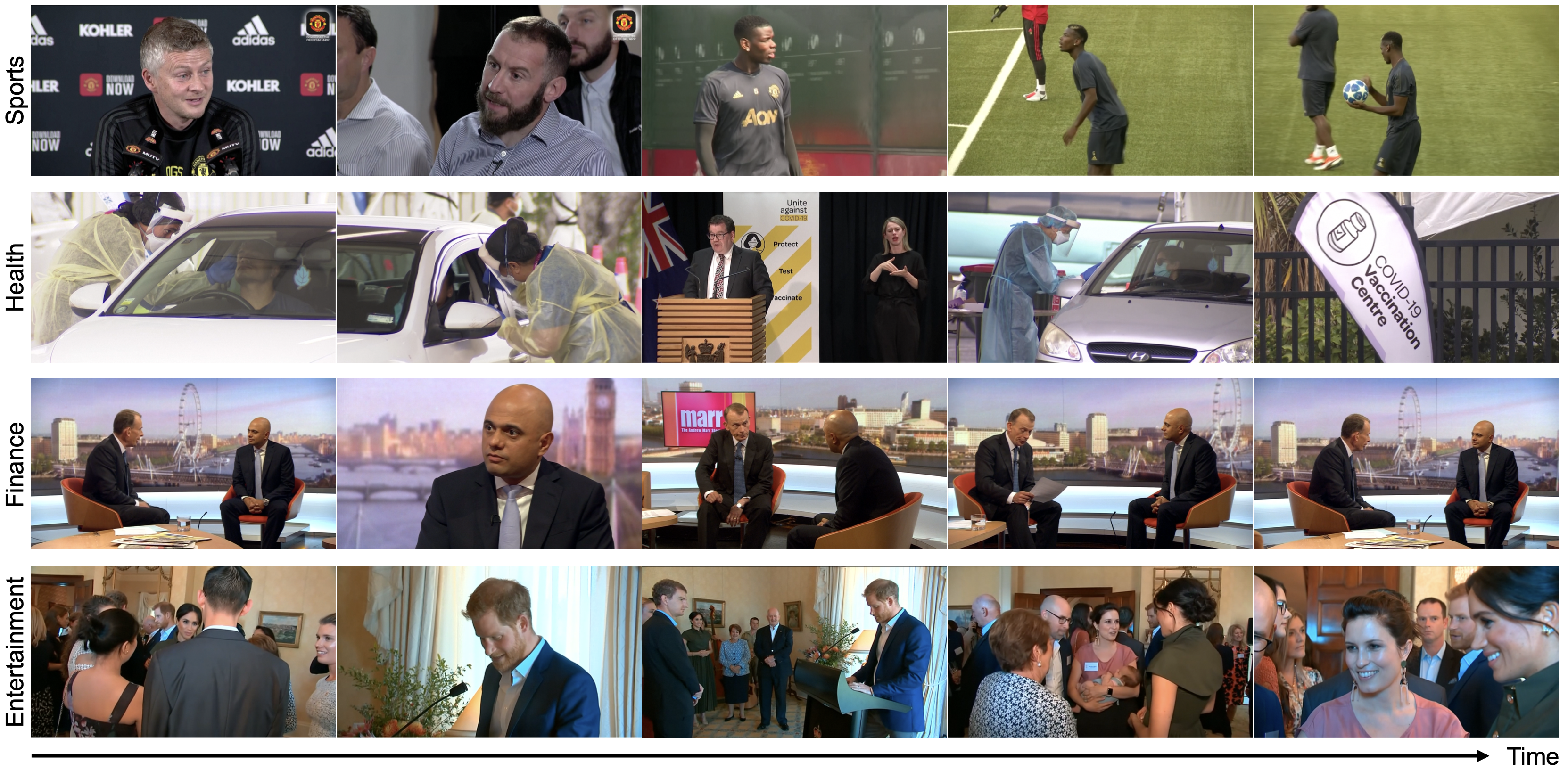}
    \caption{\textbf{Dataset Examples.} Four example videos from the \datasetname dataset. Frames are shown from four news categories with time increasing from left to right. Note the diversity of the events and visual information in each video clip. The stories covered are (i) Pogba an injury doubt for United ahead of Premier League clash with Arsenal; (ii) New Zealand's COVID-19 cases hit record for second time this week; (iii) Britain will have "good set" of Brexit policies ready for June EU summit - minister; (iv) `Couldn't think of a better place' to announce baby: Prince Harry in Australia.}
    \label{fig:DataExample}
\end{figure}

The challenge of understanding video and language simultaneously remains an active area of research in machine learning~\cite{BMT_Iashin_2020, MDVC_Iashin_2020, song2021paragraph, chou2023implicit, chou2022semi, liu2021video}.
This task requires the ability to analyze and interpret both visual and linguistic information; 
several video understanding sub-tasks have been proposed,
such as video captioning~\cite{anderson2018bottom, sharma2018conceptual, seo2022end, MDVC_Iashin_2020, BMT_Iashin_2020} (i.e., given a video, provide a caption describing the video), video question answering~\cite{yu2018joint} (given a video and a question, provide the correct answer to the question), and text-video retrieval~\cite{liu2022ts2net} (i.e., given a dataset of videos and sentences describing the videos, match the corresponding text-video pairs).
While recent years have witnessed significant progress in the development of algorithms that can jointly learn from video and language data, many open challenges still exist, including the lack of high-quality benchmark datasets, the difficulty of modeling complex temporal dependencies~\cite{han2022temporal}, and the need for more effective and interpretable models. Therefore, the development of diverse and high-quality datasets must be prioritized to advance the state-of-the-art in video and language understanding.
Several datasets~\cite{zhou2018towards, rohrbach2015dataset, sigurdsson2016hollywood, chen2011collecting, xu2016msr, zeng2016title, krishna2017dense, gella2018dataset, miech19howto100m} have been released to benchmark models on multiple video understanding tasks such as video classification~\cite{abu2016youtube,monfort2019moments, islam2022long, uchiyama2023visually}, video captioning~\cite{krishna2017dense, sharma2018conceptual, seo2022end, MDVC_Iashin_2020, BMT_Iashin_2020}, video paragraph generation~\cite{song2021paragraph, liu2021video}, and video-text retrieval~\cite{liu2022ts2net, DRLTVR2022, luo2022clip4clip}. The majority of these existing video benchmarks are either web-scrapped from the internet~\cite{abu2016youtube, krishna2017dense, zeng2016title}, manually recorded and labeled by paid laypersons~\cite{krishna2017dense, xu2016msr} (e.g., Amazon Turk-like services), or relatively small in size~\cite{rohrbach2015dataset, chen2011collecting}. Moreover, the level of description of existing video-language datasets remains relatively low and focuses on describing events that visibly occur in the video. 

Most of the approaches~\cite{islam2022long, uchiyama2023visually, tong2022videomae, zhang2022actionformer} focus on the visual part to understand the content of the video. However, humans leverage visual, audio, linguistic, and even contextual information to understand videos at a higher level.
For example, when watching a video about a natural disaster, a human understands the location of the event, the impact on surrounding areas, a sense of the level of damage, and other phenomena which are not visible by observing the pixels of the video. Despite this, there is a lack of video datasets that contain such abstract levels of textual annotations for video. Such a dataset would be beneficial as it would provide new technical challenges at higher levels of abstraction which would require the models to employ richer contextual information to solve the task at hand.

To spur such research,
we propose a new video-language multimodal dataset: \textit{the Reuters News agency Video-Language dataset shortened to "\datasetname" dataset}, for a variety of video and video-language understanding tasks.
\datasetname contains news events videos, audio, and texts with professionally curated and labeled from around the world. It is collected by journalists from dozens of countries and contains rich and consistent labels generated by professionals in the news industry. To demonstrate the utility of the dataset, we benchmark various state-of-the-art algorithms for four different video-language understanding tasks and uncover interesting findings with respect to the type of challenges deep networks have with the dataset. Finally, we suggest future directions for addressing these challenges as well as open problems in the domain of video-language understanding.

This technical report can be summarized as follows: (i) We present an analysis on a new in-house dataset, \datasetname: $1,974$ news story videos from $2018/04$ to $2021/12$ filmed and annotated by professional journalists and editors from the Reuters News agency, with the purpose of understanding current challenges in long-form news video understanding. The dataset contains videos from $200$+ different global locations and spans seven subject categories: \textit{disaster, finance, entertainment, health, politics, sports, and miscellaneous}. The language annotations of the dataset span various levels of abstraction (e.g., both visually observable actions and contextual background information) which include \textit{title caption, visual video description, high-level story description, keywords, and location}. (ii) We then perform a comprehensive analysis 
    of \datasetname by comparing the statistics with other datasets and by evaluating several state-of-the-art deep models for four open problems in video-language understanding using the dataset: (1) video captioning, (2) video paragraph generation, (3) open-world video keyword generation, and (4) video-text retrieval. Our comprehensive experimental results suggest that video understanding of world events and news stories is a challenging domain.

\section{Related Work}

\begin{table}[!t]
  \caption{\textbf{Comparison of \datasetname with other video language datasets.} \#videos: number of videos in the corresponding dataset. Avg. len: average length of videos in seconds. Total len: Total length of videos in hours. Cap: captions. KeyW: keywords. C. Cap: closed caption.}
  \label{tab:ComparisonDataset}
  \centering
  \small
  \resizebox{1.0\textwidth}{!}{
  \begin{tabular}{c|ccccc|cccc}
    \toprule
    \multicolumn{1}{}{} & \multicolumn{5}{|c}{Video
    Statistics} & \multicolumn{4}{|c}{Annotations}\\
    Dataset & Source & Content & \#videos & Avg. len & Total len & Cap. & Story & KeyW & C. Cap.\\
    \hline
    \hline
    YouCook II~\cite{zhou2018towards} & YouTube &Cooking & 2,000 & 316 sec & 176 hrs & \checkmark & \xmark & \xmark & \xmark \\
    MPII-MD~\cite{rohrbach2015dataset} &Films & Movie Scenes& 68,337 & 3.9 sec & 73.6 hrs & \checkmark & \xmark & \xmark & \xmark \\
    Charades~\cite{sigurdsson2016hollywood} & Mech. Turk & Home Videos & 9,848 & 30 sec & 82 hrs & \checkmark & \xmark & \xmark & \xmark \\
    MSVD~\cite{chen2011collecting} & YouTube & Open & 1,970 & 10 sec & 5.3 hrs & \checkmark & \xmark & \xmark & \xmark\\
    MSR-VTT~\cite{xu2016msr} & YouTube & Open & 7,180 & 20 sec & 41.2 hrs & \checkmark & \xmark & \xmark & \xmark \\
    VTW~\cite{zeng2016title} & YouTube & Open & 18,100 & 90 sec & 213.2 hrs & \checkmark & \xmark & \xmark & \xmark \\
    ANet Caption~\cite{krishna2017dense} & YouTube & Open & 20,000 & 180 sec & 849 hrs & \checkmark & \xmark & \xmark & \xmark \\
    Howto100M~\cite{miech19howto100m} & YouTube & Instructional & 1.2M & 396 sec &  134,472 hrs & \checkmark & \xmark & \xmark & \xmark \\
    VideoStory~\cite{gella2018dataset} & Social Media & Story Telling & 20,147 & 70 sec & 396 hrs & \checkmark & \checkmark & \xmark & \xmark \\
    \hline
    \datasetname &  Journalists &  News & 1,974 & 91.2 sec & 50 hrs & \checkmark & \checkmark & \checkmark & \checkmark\\
    \bottomrule
  \end{tabular}}
\end{table}

\textbf{Video Description.} 
The video description aims to generate sentences that describe the video automatically. It includes but is not limited to the tasks such as video captioning and video paragraph generation. Video captioning first dealt with SVO (Subject, Object, Verb) tuples-based models~\cite{koller1991algorithmic, das2013thousand, kojima2002natural}. These approaches leverage objects and actions in the video and fit them into pre-defined sentence templates. With the success of deep learning models, later approaches~\cite{venugopalan2015sequence, yao2015describing} frame the captioning task as one of the machine translation variants. In detail, Convolutional Neural Networks (CNNs)~\cite{krizhevsky2017imagenet, carreira2017quo, xie2017rethinking} are used as the encoders to model the visual data, and Recurrent Neural Networks (RNNs) are used as the decoders to generate the sentences. Chen \emph{et al.}~\cite{chen2019motion} use attention strategies to aggregate the entire video temporally to capture motion dynamics better. To alleviate the computational challenge of using expensive 3D CNNs applied to dense frame inputs, Seo \emph{et al.}~\cite{seo2022end} use a Transformer-based encoder applied to raw pixels, sampled at a coarse rate to capture long context. To give models more information to understand the videos, some works~\cite{BMT_Iashin_2020, MDVC_Iashin_2020} take not only visual input but also utilize multi-modal inputs such as audio and speech.
Video paragraph generation is generally viewed as a harder task compared with video captioning. It requires the model to handle long-term dependencies in the video and summarize the entire video content by generating multiple sentences. Liu \emph{et al.}~\cite{liu2021video} propose a framework by modeling this task as a text summarization problem. The model first generates several sentence-level captions and then summarizes them into a paragraph. On the other hand, Song \emph{et al.}~\cite{song2021paragraph} propose a one-stage framework for the video paragraph generation task by leveraging dynamic video memory and directly generating a paragraph.

\textbf{Text-Video Retrieval.} Text-Video retrieval aims to retrieve the most similar video given a textual query (or vice-versa). Numerous approaches to this problem revolve around offline feature extraction~\cite{yu2018joint,chen2020fine,gabeur2020multi,dzabraev2021mdmmt}. However current methods mainly train deep models in an end-to-end manner~\cite{lei2021less,fang2021clip2video,liu2022ts2net,luo2022clip4clip,DRLTVR2022,bain2022clip}. More recently, multiple papers~\cite{fang2021clip2video,luo2022clip4clip,bain2022clip} have proposed to leverage the CLIP~\cite{radford2021learning} model as a backbone for text-video retrieval due to the high-level semantics encoded, and related to language, contained in the CLIP model. Apart from CLIP-based approaches, TS2-Net~\cite{liu2022ts2net} proposes a token selection module that dynamically adjusts the selection criteria across temporal and spatial dimensions of the input video, while DRL~\cite{DRLTVR2022} exhaustively compares all input tokens and simultaneously learns to minimize the temporal redundancy of the sampled representation. 

\textbf{Video Language Datasets.} 
The datasets for video-language tasks are the keys to the fast advancement of the research area. We compare the statistics and annotations of the existing datasets~\cite{zhou2018towards, rohrbach2015dataset, sigurdsson2016hollywood, chen2011collecting, xu2016msr, zeng2016title, krishna2017dense, gella2018dataset, miech19howto100m} in Table~\ref{tab:ComparisonDataset}. The datasets are categorized into four main classes: Cooking, Movies, YouTube, and Social Media. In contrast, our proposed \datasetname dataset focuses on the News domain which is different from existing video-language datasets. Most of these existing datasets are collected by scrapping the internet and are automatically labelled or labelled by crowdsourcing. Furthermore, they mostly contain single-sentence descriptions for one type of annotation, such as captions or stories. Ours, on the other hand, has a variety of annotations, such as captions, stories, keywords, etc. For example, the VideoStory dataset~\cite{gella2018dataset} includes multi-sentence descriptions of events visibly occurring in the video. Contrastingly, the story in \datasetname contains the background information (e.g., location, date, and other stakeholders than the involved parties within the video) that cannot be directly inferred from the video frames and ultimately requires combining external knowledge with the available visual information.

\section{Reuters Video-Language News Dataset}
\begin{figure}[!t]
    \centering    
    \includegraphics[width=1.0\textwidth]{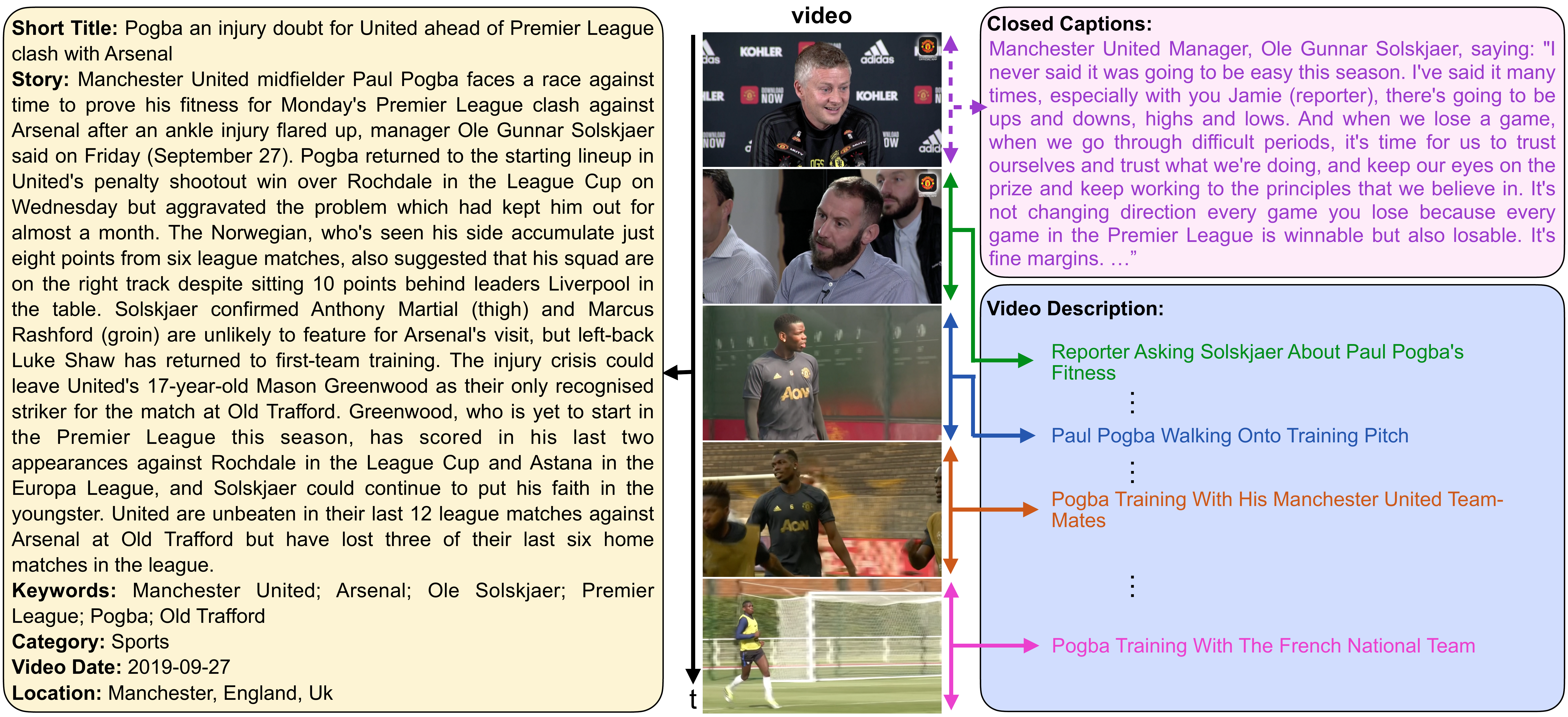}
     \vspace{-0.3cm}
    \caption{\textbf{Dataset details.} Here is an example from the \datasetname dataset. \datasetname contains the video and corresponding metadata, including short title, story, keywords, category, video description for video clips, closed caption, video date, and location. More details about the metadata information are in 
    Section~\ref{subsec:LabelsMetadata}.}
    \label{fig:DatasetDetails}
\end{figure}
The Reuters Video-Language News Dataset (\datasetname) is a dataset consisting of $1,974$ long-form news videos 
with an average video length of $91.2$ seconds. Each video reports on a single specific news story and may contain action shots of the event itself, interviews with people associated with the event, footage from nearby areas, and more. Figure~\ref{fig:DataExample} shows several examples from the dataset. The videos in \datasetname are professionally produced, annotated, and fully licensed. The dataset has coverage from $200$+ global locations and in $16$ languages. Reuters is the world’s largest international multimedia news agency and the leading provider of real-time news and intelligence. Moreover, a recent study by the Economist~\cite{economist_study2019} showed that Reuters has the highest `accuracy score' of all the publications included, and tracks at the center on `bias score', showing neither left- nor right-wing bias, making suitable for both research and application purposes.

\begin{table}[!t]
  \caption{\textbf{Dataset Statistics.} The first section is the video statistics. In order to keep the diversity of the dataset, we make sure each category has at least $200$ videos. The second and third sections are the caption and story statistics, respectively. We show the average length of the text part and the percentage of the tokens covered by Glove-6B~\cite{pennington2014glove}. The final section is the dataset splits. We split the dataset to train/val/test with $70/15/15$ ratio.}
  \label{tab:DatasetStatistics}
  \centering
  \small
  \begin{tabular}{c|c|c|ccccccc}
    \toprule
      & & Full & Entertain. & Misc & Disaster & Finance & Health & Politics & Sports \\
    \hline
    \hline
    \multirow{2}{*}{\rotatebox[origin=c|]{90}{Stat.}} & \#videos & 1,974 & 200 & 224 & 200 & 325 & 346 & 479 & 200 \\
    & Avg. Length (s) & 91.2 & 114.6 & 85.6 & 95.4 & 87.2 & 90.4 & 83.3 & 96.6 \\
    \hline
    \multirow{2}{*}{\rotatebox[origin=c|]{90}{Cap.}} & Avg. Length & 11.3 & 11.7 & 11.2 & 10.6 & 11.9 & 11.1 & 11.6 & 10.8 \\
    & \% in Glove~\cite{pennington2014glove} & 91.7 & 94.2 & 95.2 & 96.5 & 94.9 & 94.2 & 92.7 & 93.4 \\ 
    \hline
    \multirow{2}{*}{\rotatebox[origin=c|]{90}{Story}} & Avg. Length & 178.0 & 185.7 & 169.1 & 155.9 & 175.6 & 183.4 & 173.7 & 207.1 \\
    & \% in Glove~\cite{pennington2014glove}  & 85.1 & 93.0 & 93.1 & 93.2 & 91.7&  91.0 & 91.5 & 91.2 \\ 
    \hline
    \hline
    \multirow{3}{*}{\rotatebox[origin=c|]{90}{Splits}} & Train & 1,382 & 142 & 153 & 134 & 227 & 253 & 342 & 131 \\
    & Validation & 296 & 26 & 34 & 33 & 52 & 45 & 72 & 34 \\
    & Test & 296 & 32 & 37 & 33 & 46 & 48 & 65 & 35 \\
    \bottomrule
  \end{tabular}
\end{table}

\begin{figure}[t]
    \centering    
    \includegraphics[width=1.0\textwidth]{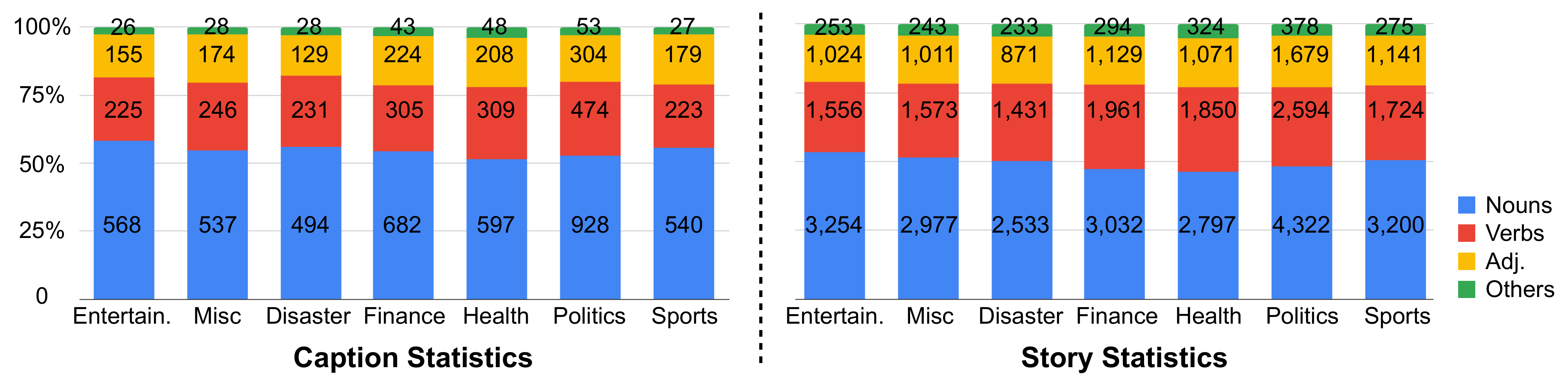}
     \vspace{-2em}
    \caption{\textbf{Text Statistics.} We show the composition of captions and stories across different categories.}
    \label{fig:LanguageStat}
\end{figure}

\subsection{Labels and Metadata}\label{subsec:LabelsMetadata}
Each video is in MP4 format and contains corresponding audio clips of reporting, interviews, etc. Each video is annotated by professionals and contains the following eight types of labels and metadata (examples are shown in Fig.~\ref{fig:DatasetDetails}). (1) \textbf{Short Title}: a one-sentence caption describing the news story covered in the video. (2) \textbf{Story}: {A long multi-sentence description of the entire story in detail which may contain contents that are not visually observable in the video, such as references to places, names, and events.} (3) \textbf{Keywords}: Each video is labeled with multiple keywords used for identifying and tagging similar videos. (4) \textbf{News Category}: A manually labeled categorization (there are seven categories total) of the topic covered in the story (e.g., sports). (5) \textbf{Video Description}: Short phrases describing the visual phenomenon in the video grounded to each camera shot. (6) \textbf{Closed Captions}: Manually transcribed captions corresponding to speaking in the audio. (7) \textbf{Video Date}. (8) \textbf{Location}. 

The \datasetname dataset contains rich semantic labels and metadata for a diverse range of topics from around the world (see Sec.~\ref{sec:statistics} for the quantitative statistics of \datasetname). In this paper, our aim is to outline several challenging tasks on news dataset. However, we note that additional tasks will be possible with the dataset in its current state (e.g., as ground truth for text-to-video generation), or alternatively, if additional annotations are produced (e.g., temporal annotations for video grounding). A deeper discussion of these tasks and future directions are given in Secs.~\ref{sec:experiments} and~\ref{sec:conclusion}.

\subsection{Labeling Guidelines}\label{subsec:LabelGuidelines}
There are several differences in the labeling guidelines given to Reuters journalists (who have years or even decades of experience) and labeling strategies such as Mechanical Turk. First, journalists are given high-level guidelines when labeling videos, rather than fine-grained instructions on exactly what to label and what to ignore. The general instructions given to all journalists when labeling the videos are to prioritize the following factors in their labeling: objectivity, bias, truth, standards of integrity, the presentation of information, and details of how to maintain quality in specific situations or with specific kinds of reporting.
Indeed, a complete set of rules that covers all situations that can occur on a daily basis is impossible to enumerate. It is worth re-emphasizing that Reuters journalists who create video content usually have years, and in some cases decades of experience providing consistent labels and annotations on video. Moreover, all Reuters journalists complete internal training to comply with those guidelines before serving in a news group under editorial staff. All published videos are also subject editorial review before publication to ensure compliance with the above guidelines. The videos are then distributed to thousands of news organizations around the planet. While inconsistencies or errors in labeling or metadata do sometimes occur in fast-breaking complex news narratives, Reuters dynamically iterates on the dataset labels, as this distribution and real world use results in any errors being reported back to us by our partners and fixed very quickly.



\subsection{Dataset Statistics}\label{sec:statistics}
This section provides a comprehensive description of the video dataset and its associated splits for training, validation, and testing. We adopt a balanced approach in selecting samples from each category to ensure that models trained on it can generalize well to unseen data. The training, validation, and testing splits are made up of 70\% (1,382 videos), 15\% (296 videos), and 15\%, respectively. We select videos such that all categories contain a minimum of 200 videos in order to reduce the negative impacts of categorical bias.

Table~\ref{tab:DatasetStatistics} and Fig.~\ref{fig:LanguageStat} provide further details about the number of videos in each category and their average length. Each video in the dataset is annotated by a caption, a long title, and a set of keywords that can be used for more granular analysis. Furthermore, it is noteworthy that the distribution of the data is fairly uniform, which assists machine learning models and prevents data imbalances from skewing the results of the model. 
\section{Experiments}\label{sec:experiments}
In this section, we provide benchmarks for a meaningful sampling of the video-language tasks possible using \datasetname. For this purpose, we present experimental results for four varied tasks on the dataset: (i) video captioning (Sec.~\ref{sec:video_caption}), (ii) video paragraph generation (Sec.~\ref{sec:paragraph}), (iii) open-world keyword generation (Sec.~\ref{sec:keyword}, and (iv) text-video retrieval (Sec.~\ref{sec:retrieval}). Note that we specifically chose tasks such that they vary in their difficulty (e.g., video captioning vs. paragraph generation), output (e.g., generative vs. discriminative), and input (e.g., text vs. video vs. audio). 

\subsection{Task 1: Video Captioning}\label{sec:video_caption}
Given a video, the video captioning task aims to generate a one-sentence description of the video. We train and evaluate two baselines, BMT~\cite{BMT_Iashin_2020} and MDVC~\cite{MDVC_Iashin_2020} on \datasetname. We take the \textbf{Short Title} as our ground truth caption (see Fig.~\ref{fig:DatasetDetails} for an example). For both models, we follow the same training settings as in the original papers except we change the dictionary size to $3,841$ and use the ground truth proposals. We follow the original papers and train the models with two different settings: (i) visual-only and (ii) visual+audio.

\begin{table}
  \caption{\textbf{Quantitative Results for Video Captioning.} The video captioning task is trained on two baselines, MDVC~\cite{MDVC_Iashin_2020} and BMT~\cite{BMT_Iashin_2020}, with two different settings, w/o and w/ audio input. We evaluate metrics with two traditional metrics, BLEU@n~\cite{papineni2002bleu}, METEOR~\cite{denkowski2014meteor}.}
  \label{tab:VideoCaptioning}
  \centering
  \small
  \begin{tabular}{c|c|cccc}
    \toprule
    & Model & BLEU3$\uparrow$ & BLEU4$\uparrow$ & METEOR$\uparrow$ \\
    \hline
    \hline
    \multirow{2}{*}{Visual-only} 
     & MDVC~\cite{MDVC_Iashin_2020} & 19.56 & 12.11 & 7.44 \\
     & BMT~\cite{BMT_Iashin_2020} & 19.82 & 12.30 & 6.79 \\
    \hline
    \multirow{2}{*}{Visual + Audio} 
     & MDVC~\cite{MDVC_Iashin_2020} & 20.06 & 12.55 & 8.00 \\
     & BMT~\cite{BMT_Iashin_2020} & 20.14 & 12.65 & 7.75 \\
    \bottomrule
  \end{tabular}
\end{table}

\begin{figure}[t]
    \centering    
    \includegraphics[width=1.0\textwidth]{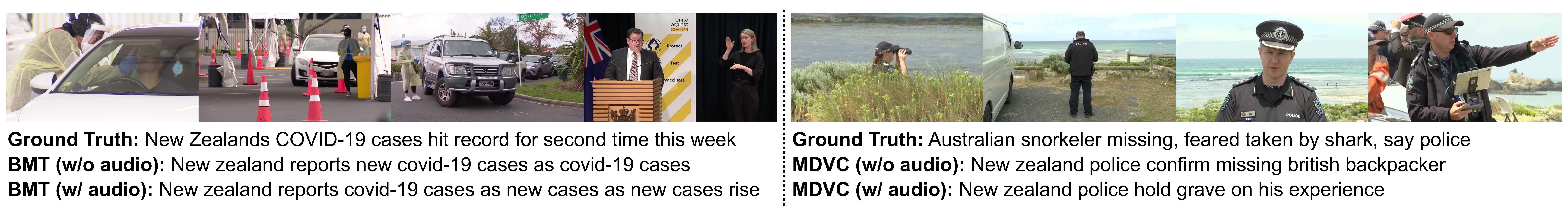}
    \caption{\textbf{Qualitative Results for Video Captioning.} We show two qualitative examples from each model and each setting. As shown in the figure, BMT~\cite{BMT_Iashin_2020} can generate more related captions compared with others. Anecdotally, one can also see that training with audio input describes the video more thoroughly, as \datasetname contains interviews and press conferences. 
    }
\label{fig:qualitative_captioning}
\end{figure}

 The results for both models and settings are shown in Table~\ref{tab:VideoCaptioning} where the performance of both models is evaluated using BLEU@n~\cite{papineni2002bleu} score and METEOR~\cite{denkowski2014meteor} score. We also visualize the results from different settings in Fig.~\ref{fig:qualitative_captioning}. The results show that using visual and audio data performs better than only using visual data only. It is expected as \datasetname contains many clips of audio-centric reporting events, such as press conferences and interviews.

\subsection{Task 2: Video Paragraph Generation}\label{sec:paragraph}
Given a video, the goal of the video paragraph generation task is to generate a multi-sentence paragraph that plausibly describes the video. Compared to the video captioning task (Sec.~\ref{sec:video_caption}), this task produces longer descriptions that offer a more detailed description over longer time periods. In this task, we train and evaluate the state-of-the-art video paragraph generation model by Song et al.~\cite{song2021paragraph} on \datasetname. The input of the model is the entire video, and we take the \textbf{Story} (see Fig.~\ref{fig:DatasetDetails} for an example) as the target of the generated paragraph. A key component of the model is the keyframe selection module, which learns to subsample important frames from the video to reduce computation. We follow the same training and evaluation settings as in the original paper.

\begin{figure}[t]
    \centering    
    \includegraphics[width=1.0\textwidth]{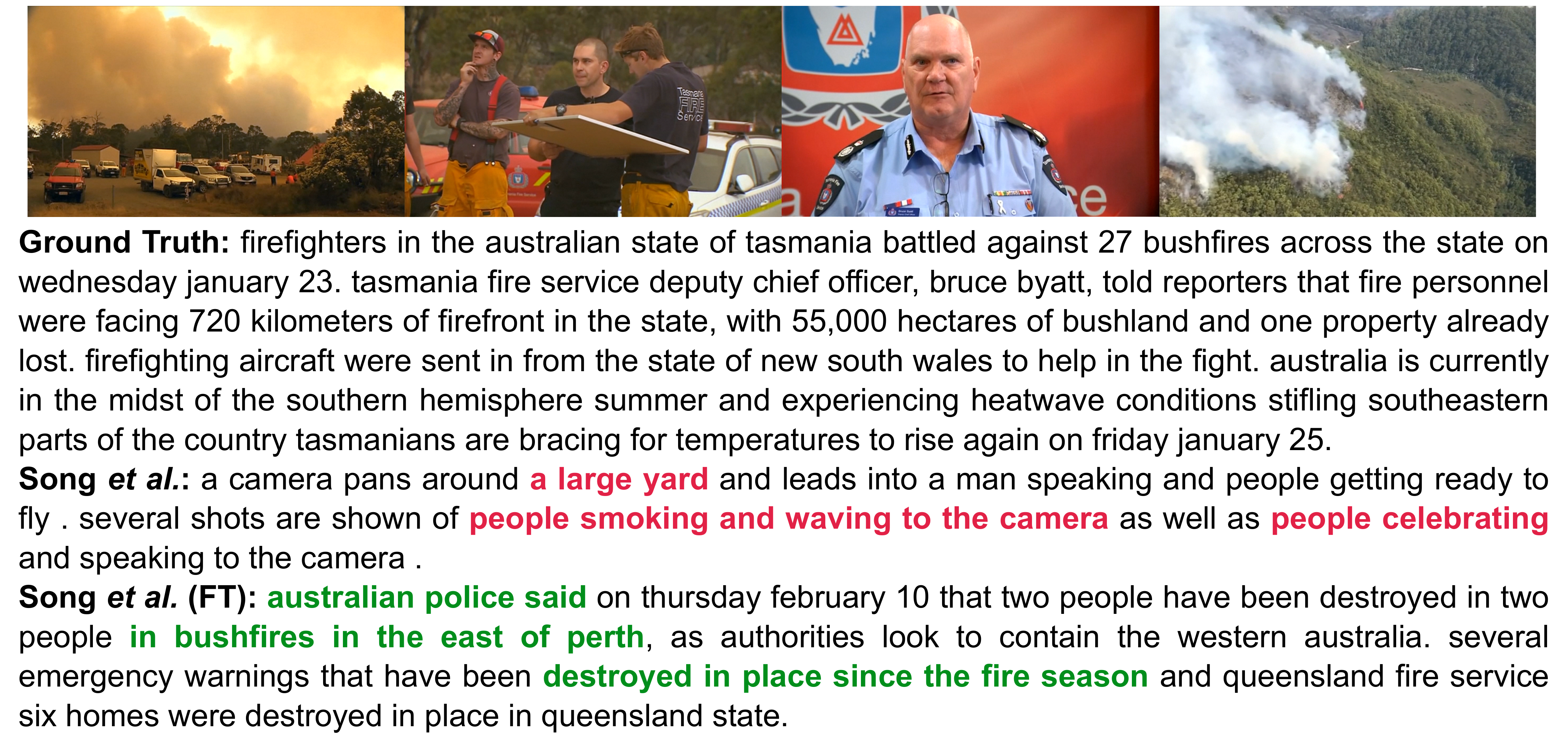}
    \caption{\textbf{Qualitative Results for Video Paragraph Generation.} We visualize the video and output paragraph from the baseline~\cite{song2021paragraph} and highlight the related content in green.}
    \label{fig:qualitative_paragraph}
\end{figure}

The paragraph generation results are shown in Table~\ref{tab:VideoParagraph}. We first evaluate the performance using accuracy (BLEU@n and METEOR score), and diversity (n-gram diversity~\cite{shetty2017speaking}: Div@n, and n-gram repetition~\cite{xiong2018move}: Rep@n) metrics. Interestingly, while the diversity is strong, the accuracy (i.e., BLEU@4 and METEOR score) for both models is quite low. 
Although finetuning increases the accuracy slightly, the scores are not high enough to draw any meaningful conclusions, e.g., the BLEU@4 is only $0.7$. 
We hypothesize that, due to the abstract and high-level nature of this dataset, these accuracy metrics are incapable of measuring adequate \textit{semantic} similarity between the prediction and target paragraphs. 
Towards a solution to this problem, we evaluate the models based on semantic similarity metrics (Sentence-Similarity~\cite{reimers-2019-sentence-bert} and Bert F-score~\cite{bert-score} ($F_{BERT}$)). These metrics are based on large language model encodings of the prediction and target paragraphs. 
As expected, these model-based similarity metrics demonstrate that finetuning improves the model output significantly.
These results agree with the qualitative results observed in Fig.~\ref{fig:qualitative_paragraph}, where the finetuned model makes less mistakes and captures more appropriate context than the non-finetuned model.

\begin{table}
  \caption{\textbf{Quantitative Results for Video Paragraph Generation.} We train the state-of-the-art video paragraph generation model~\cite{song2021paragraph} on \datasetname and evaluate on three domains, accuracy, diversity, and similarity. B@4, M, $P_B$, $R_B$, and $F_B$ are BLEU@4, METEOR, $P_{BERT}$, $R_{BERT}$, and $F_{BERT}$ respectively. FT denotes the model is finetuned on \datasetname.
  }
  \label{tab:VideoParagraph}
  \centering
  \small
    \begin{tabular}{c|cc|ccc|cc}
    \toprule
    \multicolumn{1}{}{} & \multicolumn{2}{|c}{Accuracy} & \multicolumn{3}{|c}{Diversity} & \multicolumn{2}{|c}{Similarity} \\
    Model & B@4$\uparrow$ & M$\uparrow$ & Div@1$\uparrow$ & Div@2$\uparrow$ & Rep@4$\downarrow$ & Sen-Sim$\uparrow$ & $F_{B}\uparrow$\\
    \hline
    \hline
    Song {\em et al.}~\cite{song2021paragraph} & 0.01 & 2.7 & 69.7 & 85.5 & 2.2 & 4.9 & 80.1 \\
    Song {\em et al.}~\cite{song2021paragraph} (FT) & 0.7 & 5.4 & 73.8 & 87.1 & 2.1 & \textbf{33.5} & 81.1 \\

    \bottomrule
  \end{tabular}
\end{table}

\subsection{Task 3: Open World Keyword Generation}\label{sec:keyword}

Given a video, the task of keyword generation is to tag the video with keywords that describe the content of the video. This is a difficult task because (i) the task has an open vocabulary and (ii) there is a potentially unbounded number of keywords (however, in practice, the upper limit of keywords in the dataset is around seven). We implement two baselines for this task. For the first baseline, we treat the task as a supervised classification problem and set the number of output logits of a classification model to the total number of keywords. However, note that this baseline is bounded above due to the fact that the validation and test sets contain keywords that are not found in the training set. For the model architectures, we modify 2D ResNets~\cite{he2016deep} to take 12 uniformly sampled frames from the video and then apply spatio-temporal global average pooling to the features before the linear layer. The second baseline is to use the CLIP model~\cite{radford2021learning} in a zero-shot setting. Following the original paper, we query the CLIP model with the phrase ``\textit{a photo of \{\}}'' for each frame in a video, and average the score across all frames. We then take all keywords whose score sits above the $95^{th}$ percentile of the maximum logit value. 

The keyword generation results are presented in Table~\ref{tab:keyword}. As expected, the resulting F1 scores are quite low for all baselines. Interestingly, the supervised results outperform the zero-shot baselines, even though the performance of the supervised setting is bounded above based on the intersection of the keywords between the training and validation set. We observe that ResNet18 slightly performs ResNet50 and that both CLIP baselines perform comparably to each other. The results suggest that these simple baselines are not sufficient for solving the task of open-vocabulary keyword generation. 

\begin{table}
  \caption{\textbf{Open World Keyword Generation.} We evaluate zero-shot and supervised baselines for the task of open-world keyword generation on \datasetname. This task poses significant challenges given the rare occurances of many keywords.}
  \label{tab:keyword}
  \centering
  \small
  \begin{tabular}{c|c|cccc}
    \toprule
    Model & Zero-Shot & Recall & Precision & F1 \\
    \hline
    \hline
    ResNet18~\cite{he2016deep} & \xmark & 0.081 & 0.493 & 0.137 \\
    ResNet50~\cite{he2016deep} & \xmark & 0.078 & 0.370 & 0.124 \\
    \hline
    CLIP R50~\cite{radford2021learning} & \checkmark & 0.082 &  0.078 &0.076 \\
    CLIP ViT-b~\cite{radford2021learning} & \checkmark & 0.092 & 0.073 & 0.079 \\
    \bottomrule
  \end{tabular}
\end{table}

\subsection{Task 4: Video-Text Retrieval}\label{sec:retrieval}
Given a query video, the task of Video-Text Retrieval is to return the most closely associated textual description (i.e., caption) corresponding to the query (and vice-versa for Text-Video retrieval). Example applications of this task are retrieval search engines in large unlabelled video databases. For this task, we train three recent baselines: Clip4Clip~\cite{luo2022clip4clip}, DRL~\cite{DRLTVR2022}, and TS2-Net~\cite{liu2022ts2net}. While the captions in traditional datasets (e.g., MSR-VTT~\cite{xu2016msr}, ActivityNet~\cite{krishna2017dense}) focus on describing actions occurring in the video, \datasetname captions contain higher-level descriptions of the events and story contained in the video (see Fig.~\ref{fig:DatasetDetails} for examples of the Short Title). All models are trained with $12$ frames evenly spaced as input from random spatial and temporal cropped video segments. 

\begin{figure}[t]
    \centering    
    \includegraphics[width=.85\textwidth]{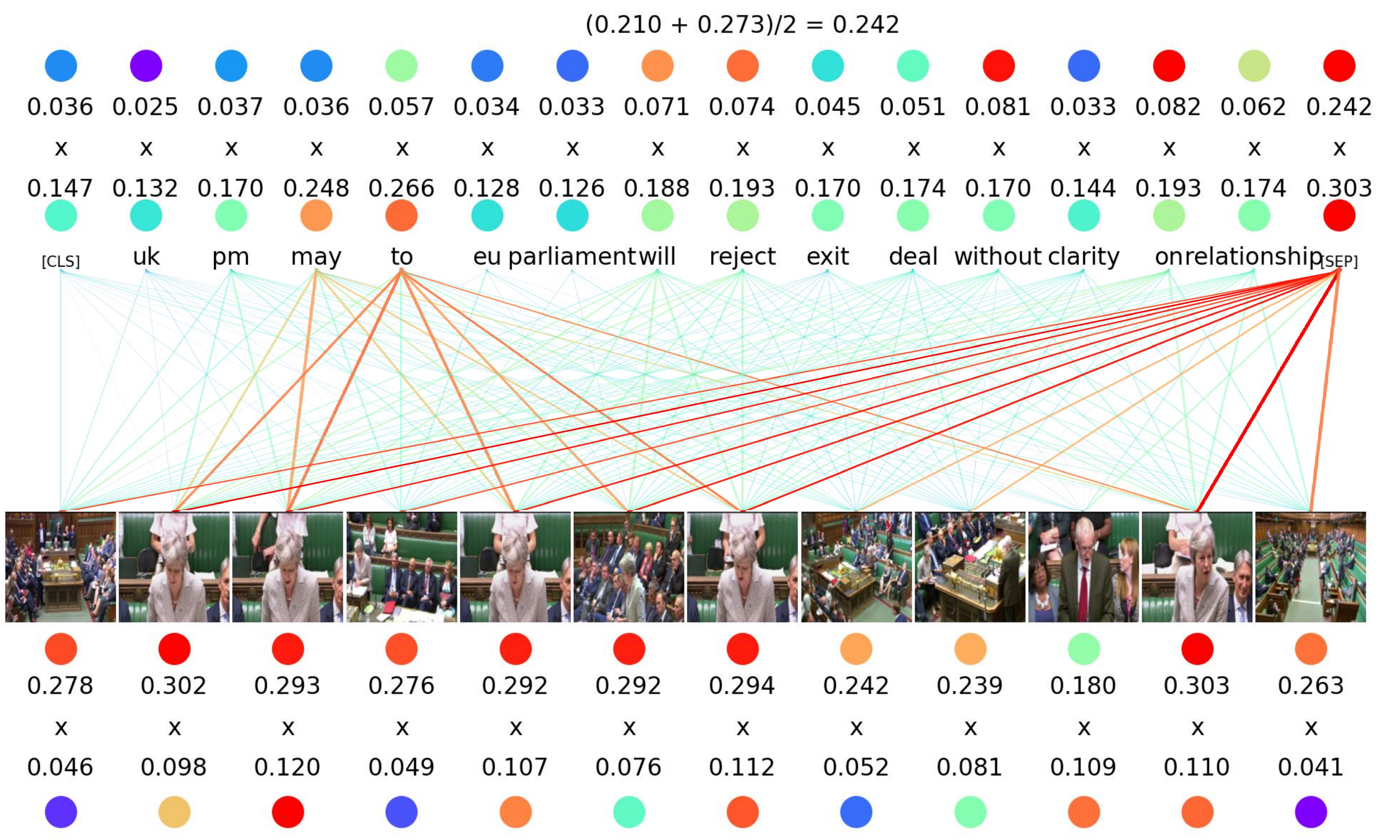}
    \caption{\textbf{Qualitative Results for Text-Video Retrieval.} We visualize the text-video token attention maps from the DRL model~\cite{DRLTVR2022} for a single video. Notice how the model attends to Theresa May over multiple frames, suggesting the model learns to recognize specific people who appear many times in the dataset.}
    \label{fig:retrieval_example}
\end{figure}

The results for the Text-to-Video (T2V) and Video-to-Text (V2T) retrieval tasks are presented in Table~\ref{tab:Retrieval}. The \datasetname dataset is a challenging dataset compared to traditional T2V retrieval benchmarks of similar size. For example, despite MSR-VTT~\cite{xu2016msr} containing almost twice as many validation videos as \datasetname (1,000 vs.\ 632), TS2-Net achieves an R@1 of $47.0\%$ on MSR-VTT compared with $51.4\%$ on \datasetname. Given the smaller number of examples to chose from during inference, we conclude that \datasetname poses a difficult challenge for the text-video retrieval task.  
Figure~\ref{fig:retrieval_example} visualizes the interpretable text-video attention mechanism from the DRL~\cite{DRLTVR2022} model. Interestingly, the model learns to recognize and attend to the UK Prime Minister, Theresa May, speaking in parliament in order to rank this video as likely belonging to the caption. This exemplifies how high-level visual phenomena, such as specific people (e.g., country leaders) and places (e.g., specific parliamentary buildings) are important for solving the \datasetname dataset. 


\begin{table}
  \caption{\textbf{Quantitative Results for Text-Video Retrieval.} We train three state-of-the-art text-video retrieval models on \datasetname. The model performances reveal that the dataset is challenging to solve and TS2-Net~\cite{liu2022ts2net} outperforms the other models.}
  \label{tab:Retrieval}
  \centering
  \small
  \begin{tabular}{c|c|cccc}
        \toprule
        & Model & R@1 & R@5 & R@10 & MnR\\
        \hline
        \hline
        \multirow{3}{*}{Text-to-Video Retrieval} & DRL~\cite{DRLTVR2022} & 36.2 & 60.4 & 72 & 15.0 \\
        & Clip4Clip~\cite{luo2022clip4clip} & 42.6 & 72.3 & 82.4 & 7.5 \\
        & TS2-Net~\cite{liu2022ts2net} & 51.4 & 80.4 & 90.5 & 5.5 \\
        \hline
        \multirow{3}{*}{Video-to-Text Retrieval} & DRL~\cite{DRLTVR2022} & 30.9 & 59 & 71.4 & 14.8 \\
        & Clip4Clip~\cite{luo2022clip4clip} & 39.5 & 72.0 & 86.1 & 6.3 \\
        & TS2-Net~\cite{liu2022ts2net} & 52 & 80.7 & 89.5 & 5.2 \\ 
        \bottomrule
    \end{tabular}
\end{table}

\section{Conclusion}\label{sec:conclusion}

This technical report presents an analysis on a large-scale video-language understanding dataset, the Reuters Video-Language News (\datasetname) dataset. \datasetname consists of $1,974$ news-oriented videos covering seven diverse categories and are collected and labeled by professionals in the news industry. Each video contains a video caption (short title), a story that describes the content of the video, keywords, a video category, detailed video descriptions, closed captions, video date, and location. We benchmark the \datasetname dataset on four different video-language tasks: (i) video captioning, (ii) video paragraph generation, (iii) open-world video keywords generation, and (iv) video-text retrieval. Experiments were run for each of these tasks and revealed several interesting findings and directions for future work.For the video-captioning task (Sec.~\ref{sec:video_caption}), the audio modality can improve performance however it is likely that this can be further improved with better mixing of audio-visual information. For video paragraph generation (Sec.~\ref{sec:paragraph}), the BLEU and Meteor metrics are insufficient for measuring ‘story’ similarity. More agreeable metrics are Sentence-Similarity~\cite{reimers-2019-sentence-bert} and Bert F-score~\cite{bert-score}, which, when used, show the benefits of fine-tuning on ReutersViLNews. Keyword generation was shown to be a very difficult task for \textit{both} supervised and open-world vision language models.
Finally, visualizations showed that text-video retrieval models leverage identifying specific people who appear frequently in the dataset in order to associate the corresponding caption (or video).


\clearpage
\newpage

\section*{Acknowledgement}\label{sec:acknowledgement}
The Reuters News Agency and Thomson Reuters members and subject matter experts deserve special credit. We thank Yulia Pavlova for managing the delivery of the dataset and for her invaluable feedback and support throughout this project. We would like to thank Vector Institute for making this collaboration possible and providing academic infrastructure and computing support during all phases of this work.

{\small
\printbibliography

@inproceedings{anderson2018bottom,
  title={Bottom-up and top-down attention for image captioning and visual question answering},
  author={Anderson, Peter and He, Xiaodong and Buehler, Chris and Teney, Damien and Johnson, Mark and Gould, Stephen and Zhang, Lei},
  booktitle={CVPR},
  year={2018}
}

@inproceedings{sharma2018conceptual,
  title={Conceptual captions: A cleaned, hypernymed, image alt-text dataset for automatic image captioning},
  author={Sharma, Piyush and Ding, Nan and Goodman, Sebastian and Soricut, Radu},
  booktitle={ACL (Volume 1: Long Papers)},
  year={2018}
}

@inproceedings{seo2022end,
  title={End-to-end generative pretraining for multimodal video captioning},
  author={Seo, Paul Hongsuck and Nagrani, Arsha and Arnab, Anurag and Schmid, Cordelia},
  booktitle={CVPR},
  year={2022}
}

@article{monfort2019moments,
  title={Moments in time dataset: one million videos for event understanding},
  author={Monfort, Mathew and Andonian, Alex and Zhou, Bolei and Ramakrishnan, Kandan and Bargal, Sarah Adel and Yan, Tom and Brown, Lisa and Fan, Quanfu and Gutfreund, Dan and Vondrick, Carl and others},
  journal={IEEE transactions on pattern analysis and machine intelligence},
  volume={42},
  number={2},
  pages={502--508},
  year={2019},
  publisher={IEEE}
}

@article{economist_study2019,
  title={Reuters features in Economist study on accuracy and bias},
  author={Stephen J Adler},
  year={2019},
  publisher={Reuters}
}

@inproceedings{liu2021video,
  title={Video paragraph captioning as a text summarization task},
  author={Liu, Hui and Wan, Xiaojun},
  booktitle={Proceedings of the 59th Annual Meeting of the Association for Computational Linguistics and the 11th International Joint Conference on Natural Language Processing (Volume 2: Short Papers)},
  year={2021}
}

@inproceedings{das2013thousand,
  title={A thousand frames in just a few words: Lingual description of videos through latent topics and sparse object stitching},
  author={Das, Pradipto and Xu, Chenliang and Doell, Richard F and Corso, Jason J},
  booktitle={CVPR},
  year={2013}
}

@article{kojima2002natural,
  title={Natural language description of human activities from video images based on concept hierarchy of actions},
  author={Kojima, Atsuhiro and Tamura, Takeshi and Fukunaga, Kunio},
  journal={IJCV},
  year={2002}
}

@inproceedings{koller1991algorithmic,
  title={Algorithmic characterization of vehicle trajectories from image sequences by motion verbs.},
  author={Koller, Dieter and Heinze, Norbert and Nagel, Hans-Hellmut},
  booktitle={CVPR},
  year={1991}
}

@article{krizhevsky2017imagenet,
  title={Imagenet classification with deep convolutional neural networks},
  author={Krizhevsky, Alex and Sutskever, Ilya and Hinton, Geoffrey E},
  journal={ACM},
  year={2017}
}

@inproceedings{chen2019motion,
  title={Motion guided spatial attention for video captioning},
  author={Chen, Shaoxiang and Jiang, Yu-Gang},
  booktitle={AAAI},
  year={2019}
}

@inproceedings{carreira2017quo,
  title={Quo vadis, action recognition? a new model and the kinetics dataset},
  author={Carreira, Joao and Zisserman, Andrew},
  booktitle={CVPR},
  year={2017}
}

@article{xie2017rethinking,
  title={Rethinking spatiotemporal feature learning for video understanding},
  author={Xie, Saining and Sun, Chen and Huang, Jonathan and Tu, Zhuowen and Murphy, Kevin},
  journal={ArXiv},
  year={2017}
}

@inproceedings{venugopalan2015sequence,
  title={Sequence to sequence-video to text},
  author={Venugopalan, Subhashini and Rohrbach, Marcus and Donahue, Jeffrey and Mooney, Raymond and Darrell, Trevor and Saenko, Kate},
  booktitle={ICCV},
  year={2015}
}

@inproceedings{yao2015describing,
  title={Describing videos by exploiting temporal structure},
  author={Yao, Li and Torabi, Atousa and Cho, Kyunghyun and Ballas, Nicolas and Pal, Christopher and Larochelle, Hugo and Courville, Aaron},
  booktitle={ICCV},
  year={2015}
}

@inproceedings{bert-score,
  title={BERTScore: Evaluating Text Generation with BERT},
  author={Tianyi Zhang* and Varsha Kishore* and Felix Wu* and Kilian Q. Weinberger and Yoav Artzi},
  booktitle={ICLR},
  year={2020}
}

@InProceedings{BMT_Iashin_2020,
  title={A Better Use of Audio-Visual Cues: Dense Video Captioning with Bi-modal Transformer},
  author={Iashin, Vladimir and Rahtu, Esa},
  booktitle={BMVC},
  year={2020}
}

@inproceedings{liu2022ts2net,
      title={TS2-Net: Token Shift and Selection Transformer for Text-Video Retrieval}, 
      author={Yuqi Liu and Pengfei Xiong and Luhui Xu and Shengming Cao and Qin Jin},
      year={2022},
      booktitle={ECCV},
}

@Article{DRLTVR2022,
  author  = {Qiang Wang and Yanhao Zhang and Yun Zheng and Pan Pan and Xian-Sheng Hua},
  journal = {ArXiv},
  title   = {Disentangled Representation Learning for Text-Video Retrieval},
  year    = {2022},
}

@article{luo2022clip4clip,
  title={CLIP4Clip: An empirical study of CLIP for end to end video clip retrieval and captioning},
  author={Luo, Huaishao and Ji, Lei and Zhong, Ming and Chen, Yang and Lei, Wen and Duan, Nan and Li, Tianrui},
  journal={Neurocomputing},
  year={2022}
}

@InProceedings{MDVC_Iashin_2020,
  author = {Iashin, Vladimir and Rahtu, Esa},
  title = {Multi-Modal Dense Video Captioning},
  booktitle = {CVPR Workshops},
  year = {2020}
}

@inproceedings{song2021paragraph,
  title={Towards Diverse Paragraph Captioning for Untrimmed Videos},
  author={Song, Yuqing and Chen, Shizhe and Jin, Qin},
  booktitle={CVPR},
  year={2021}
}

@inproceedings{chen2020fine,
  title={Fine-grained video-text retrieval with hierarchical graph reasoning},
  author={Chen, Shizhe and Zhao, Yida and Jin, Qin and Wu, Qi},
  booktitle={CVPR},
  year={2020}
}

@inproceedings{yu2018joint,
  title={A joint sequence fusion model for video question answering and retrieval},
  author={Yu, Youngjae and Kim, Jongseok and Kim, Gunhee},
  booktitle={ECCV},
  year={2018}
}

@inproceedings{gabeur2020multi,
  title={Multi-modal transformer for video retrieval},
  author={Gabeur, Valentin and Sun, Chen and Alahari, Karteek and Schmid, Cordelia},
  booktitle={ECCV},
  year={2020}
}

@inproceedings{dzabraev2021mdmmt,
  title={Mdmmt: Multidomain multimodal transformer for video retrieval},
  author={Dzabraev, Maksim and Kalashnikov, Maksim and Komkov, Stepan and Petiushko, Aleksandr},
  booktitle={CVPR},
  year={2021}
}

@inproceedings{lei2021less,
  title={Less is more: Clipbert for video-and-language learning via sparse sampling},
  author={Lei, Jie and Li, Linjie and Zhou, Luowei and Gan, Zhe and Berg, Tamara L and Bansal, Mohit and Liu, Jingjing},
  booktitle={CVPR},
  year={2021}
}

@article{bain2022clip,
  title={A CLIP-Hitchhiker's Guide to Long Video Retrieval},
  author={Bain, Max and Nagrani, Arsha and Varol, G{\"u}l and Zisserman, Andrew},
  journal={ArXiv},
  year={2022}
}

@article{fang2021clip2video,
  title={Clip2video: Mastering video-text retrieval via image clip},
  author={Fang, Han and Xiong, Pengfei and Xu, Luhui and Chen, Yu},
  journal={ArXiv},
  year={2021}
}

@inproceedings{radford2021learning,
  title={Learning transferable visual models from natural language supervision},
  author={Radford, Alec and Kim, Jong Wook and Hallacy, Chris and Ramesh, Aditya and Goh, Gabriel and Agarwal, Sandhini and Sastry, Girish and Askell, Amanda and Mishkin, Pamela and Clark, Jack and others},
  booktitle={ICML},
  year={2021}
  }

@inproceedings{denkowski2014meteor,
  title={Meteor universal: Language specific translation evaluation for any target language},
  author={Denkowski, Michael and Lavie, Alon},
  booktitle={Proceedings of the ninth workshop on statistical machine translation},
  year={2014}
}

@inproceedings{papineni2002bleu,
  title={Bleu: a method for automatic evaluation of machine translation},
  author={Papineni, Kishore and Roukos, Salim and Ward, Todd and Zhu, Wei-Jing},
  booktitle={ACL},
  year={2002}
}

@inproceedings{reimers-2019-sentence-bert,
    title = "Sentence-BERT: Sentence Embeddings using Siamese BERT-Networks",
    author = "Reimers, Nils and Gurevych, Iryna",
    booktitle = "EMNLP",
    year = "2019"
}

@inproceedings{shetty2017speaking,
  title={Speaking the same language: Matching machine to human captions by adversarial training},
  author={Shetty, Rakshith and Rohrbach, Marcus and Anne Hendricks, Lisa and Fritz, Mario and Schiele, Bernt},
  booktitle={ICCV},
  year={2017}
}

@inproceedings{xiong2018move,
  title={Move forward and tell: A progressive generator of video descriptions},
  author={Xiong, Yilei and Dai, Bo and Lin, Dahua},
  booktitle={ECCV},
  year={2018}
}

@inproceedings{he2016deep,
  title={Deep residual learning for image recognition},
  author={He, Kaiming and Zhang, Xiangyu and Ren, Shaoqing and Sun, Jian},
  booktitle={CVPR},
  year={2016}
}

@inproceedings{pennington2014glove,
  author = {Jeffrey Pennington and Richard Socher and Christopher D. Manning},
  booktitle = {EMNLP},
  title = {GloVe: Global Vectors for Word Representation},
  year = {2014},
  url = {http://www.aclweb.org/anthology/D14-1162},
}

@inproceedings{chen2011collecting,
  title={Collecting highly parallel data for paraphrase evaluation},
  author={Chen, David and Dolan, William B},
  booktitle={ACL: human language technologies},
  year={2011}
}

@inproceedings{zhou2018towards,
  title={Towards automatic learning of procedures from web instructional videos},
  author={Zhou, Luowei and Xu, Chenliang and Corso, Jason},
  booktitle={AAAI},
  year={2018}
}

@inproceedings{zeng2016title,
  title={Title generation for user generated videos},
  author={Zeng, Kuo-Hao and Chen, Tseng-Hung and Niebles, Juan Carlos and Sun, Min},
  booktitle={ECCV},
  year={2016},
  organization={Springer}
}

@inproceedings{sigurdsson2016hollywood,
  title={Hollywood in homes: Crowdsourcing data collection for activity understanding},
  author={Sigurdsson, Gunnar A and Varol, G{\"u}l and Wang, Xiaolong and Farhadi, Ali and Laptev, Ivan and Gupta, Abhinav},
  booktitle={ECCV},
  year={2016},
  organization={Springer}
}

@inproceedings{xu2016msr,
  title={Msr-vtt: A large video description dataset for bridging video and language},
  author={Xu, Jun and Mei, Tao and Yao, Ting and Rui, Yong},
  booktitle={CVPR},
  year={2016}
}

@inproceedings{krishna2017dense,
  title={Dense-captioning events in videos},
  author={Krishna, Ranjay and Hata, Kenji and Ren, Frederic and Fei-Fei, Li and Carlos Niebles, Juan},
  booktitle={ICCV},
  year={2017}
}

@inproceedings{rohrbach2015dataset,
  title={A dataset for movie description},
  author={Rohrbach, Anna and Rohrbach, Marcus and Tandon, Niket and Schiele, Bernt},
  booktitle={CVPR},
  year={2015}
}

@inproceedings{gella2018dataset,
  title={A dataset for telling the stories of social media videos},
  author={Gella, Spandana and Lewis, Mike and Rohrbach, Marcus},
  booktitle={EMNLP},
  year={2018}
}

@inproceedings{chou2022semi,
  title={Semi-supervised Grounding Alignment for Multi-modal Feature Learning},
  author={Chou, Shih-Han and Fan, Zicong and Little, James J and Sigal, Leonid},
  booktitle={CRV},
  year={2022}
}

@article{chou2023implicit,
  title={Implicit and Explicit Commonsense for Multi-sentence Video Captioning},
  author={Chou, Shih-Han and Little, James J and Sigal, Leonid},
  journal={ArXiv},
  year={2023}
}

@article{abu2016youtube,
  title={Youtube-8m: A large-scale video classification benchmark},
  author={Abu-El-Haija, Sami and Kothari, Nisarg and Lee, Joonseok and Natsev, Paul and Toderici, George and Varadarajan, Balakrishnan and Vijayanarasimhan, Sudheendra},
  journal={ArXiv},
  year={2016}
}

@inproceedings{uchiyama2023visually,
  title={Visually explaining 3D-CNN predictions for video classification with an adaptive occlusion sensitivity analysis},
  author={Uchiyama, Tomoki and Sogi, Naoya and Niinuma, Koichiro and Fukui, Kazuhiro},
  booktitle={WACV},
  year={2023}
}

@inproceedings{han2022temporal,
  title={Temporal alignment networks for long-term video},
  author={Han, Tengda and Xie, Weidi and Zisserman, Andrew},
  booktitle={CVPR},
  year={2022}
}

@inproceedings{islam2022long,
  title={Long movie clip classification with state-space video models},
  author={Islam, Md Mohaiminul and Bertasius, Gedas},
  booktitle={ECCV},
  year={2022}
}

@inproceedings{tong2022videomae,
  title={Video{MAE}: Masked Autoencoders are Data-Efficient Learners for Self-Supervised Video Pre-Training},
  author={Zhan Tong and Yibing Song and Jue Wang and Limin Wang},
  booktitle={NeurIPS},
  year={2022}
}

@inproceedings{zhang2022actionformer,
  title={ActionFormer: Localizing Moments of Actions with Transformers},
  author={Zhang, Chen-Lin and Wu, Jianxin and Li, Yin},
  booktitle={ECCV},
  year={2022}
}

@inproceedings{miech19howto100m,
   title={How{T}o100{M}: {L}earning a {T}ext-{V}ideo {E}mbedding by {W}atching {H}undred {M}illion {N}arrated {V}ideo {C}lips},
   author={Miech, Antoine and Zhukov, Dimitri and Alayrac, Jean-Baptiste and Tapaswi, Makarand and Laptev, Ivan and Sivic, Josef},
   booktitle={ICCV},
   year={2019},
}
}

\end{document}